\documentclass[letterpaper, 10 pt, journal, twoside]{ieeetran} 
\IEEEoverridecommandlockouts                              

\usepackage[T1]{fontenc}
\usepackage{mathptmx} 
\usepackage{times} 
\usepackage{amsmath} 
\usepackage{amssymb}  
\usepackage{graphicx} 
\usepackage[ruled,vlined]{algorithm2e}
\usepackage{booktabs}  
\usepackage{float}
\usepackage{cite} 

\usepackage{graphicx}
\usepackage{amsmath}
\usepackage{bigstrut}
\usepackage{multirow}
\usepackage{balance}
\usepackage[hidelinks=true,bookmarks=false]{hyperref}  
\usepackage{multirow}
\usepackage{lineno}
\usepackage{comment}
\usepackage{color}
\usepackage{pbox}
\usepackage{mathtools}
\usepackage{subfigure} 
\usepackage[ruled]{algorithm2e}

\usepackage{booktabs}
\newcommand{\otoprule}{\midrule[\heavyrulewidth]}
 \usepackage{multirow}
 \newcommand{\tabincell}[2]{\begin{tabular}{@{}#1@{}}#2\end{tabular}}

 \usepackage{graphicx} 

 \usepackage{caption}
 \usepackage{cite}
\usepackage{mathptmx}
\usepackage{times}
\usepackage{amsmath}
\usepackage{amssymb} 

\title{PointTrackNet: An End-to-End Network for 3-D Object Detection and Tracking from Point Clouds
}
 
\author{Sukai Wang, Yuxiang Sun, Chengju Liu, and Ming Liu
\thanks{This work was supported in part by the National Natural Science Foundation of China (U1713211, 61673300), in part by Basic Research Project of Shanghai Science and Technology Commission (Grant No. 18DZ1200804), in part by HKUST ECE Start-up Grant from HKUST for Heterogeneous Navigation System. \textit{(corresponding author: Ming Liu)}}
\thanks{Sukai Wang, Yuxiang Sun, and Ming Liu are with The Hong Kong University of Science and Technology, Clear Water Bay, Hong Kong, China. (e-mail: swangcy@connect.ust.hk; eeyxsun@ust.hk, sun.yuxiang@outlook.com; eelium@ust.hk).}%
\thanks{Chengju Liu is with the school of Electronics and Information Engineering, Tongji University, Shanghai, China. (e-mail: liuchengju@tongji.edu.cn).}
}

\IEEEaftertitletext{\vspace{-2.5\baselineskip}}

\begin{document}
 
\maketitle

\begin{abstract}
Recent machine learning-based multi-object tracking (MOT) frameworks are becoming popular for 3-D point clouds. Most traditional tracking approaches use filters (e.g., Kalman filter or particle filter) to predict object locations in a time sequence, however, they are vulnerable to extreme motion conditions, such as sudden braking and turning. In this letter, we propose PointTrackNet, an end-to-end 3-D object detection and tracking network, to generate foreground masks, 3-D bounding boxes, and point-wise tracking association displacements for each detected object. The network merely takes as input two adjacent point-cloud frames. Experimental results on the KITTI tracking dataset show competitive results over the state-of-the-arts, especially in the irregularly and rapidly changing scenarios. 
\end{abstract}

\begin{IEEEkeywords}
Point Cloud, Multiple-object Tracking, End-to-End, Autonomous Vehicles.
\end{IEEEkeywords}

\section{Introduction}
\IEEEPARstart{M}{ultiple}-object Tracking (MOT) plays a very important role in autonomous vehicles and Advanced Driver Assistance Systems (ADAS) \cite{wang2019pointit, wang2017simultaneous}. The tracking trajectories can be used for path planning, collision avoidance, and precise pose estimation, etc. Most MOT approaches decompose the task into two stages: object detection and data association. During object detection, objects on roads can be categorized as cars, pedestrians, cyclists, and background, etc. During data association, the same objects at different time stamps are linked to form a trajectory. With the trajectory of each object, we can forecast their positions and predict potential accidents in the future.

Nowadays, various sensors, such as LiDAR, radar, camera, and GPS, have become available for driverless systems. Many methods use sensor fusion techniques for MOT, however, most of them face the challenge of how to keep the whole system working stable in the situation that any of the sensors encounters accidental damages. Besides, the system calibration error is also a major issue. So only using a single sensor could alleviate the integration issue.   

\begin{figure}[t]
	\centering
	\includegraphics[width=0.46\textwidth]{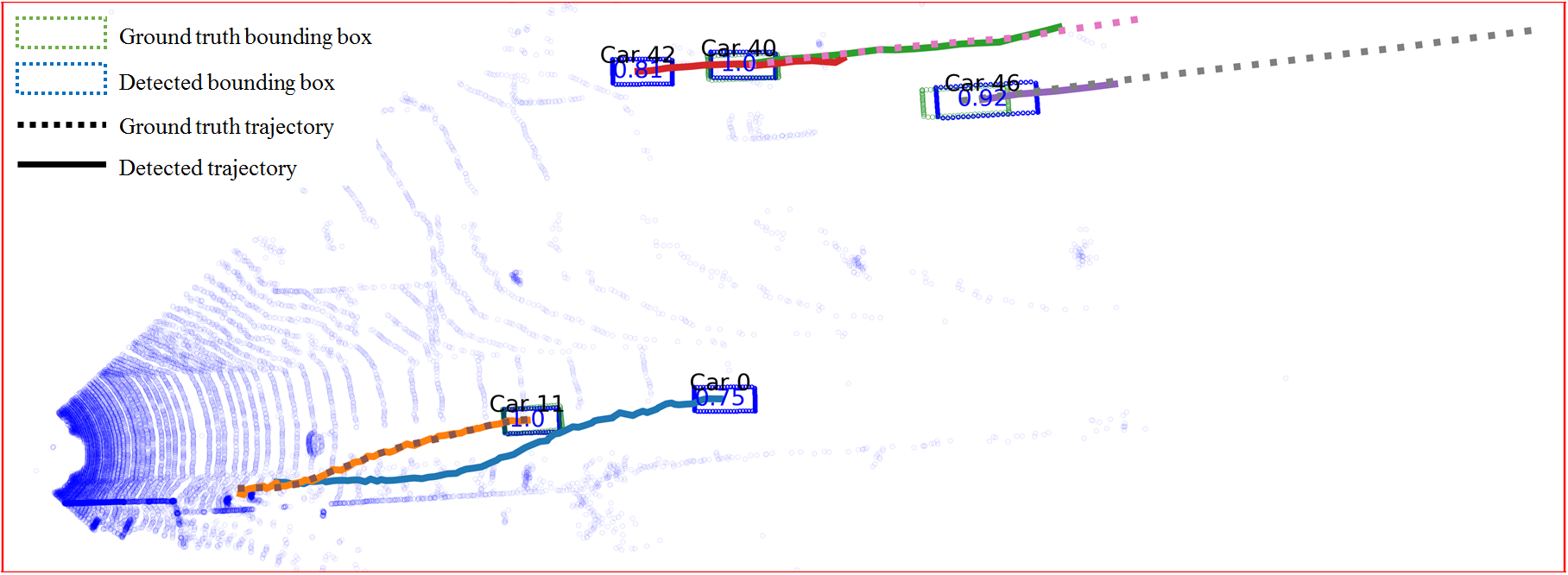}
	\includegraphics[width=0.46\textwidth]{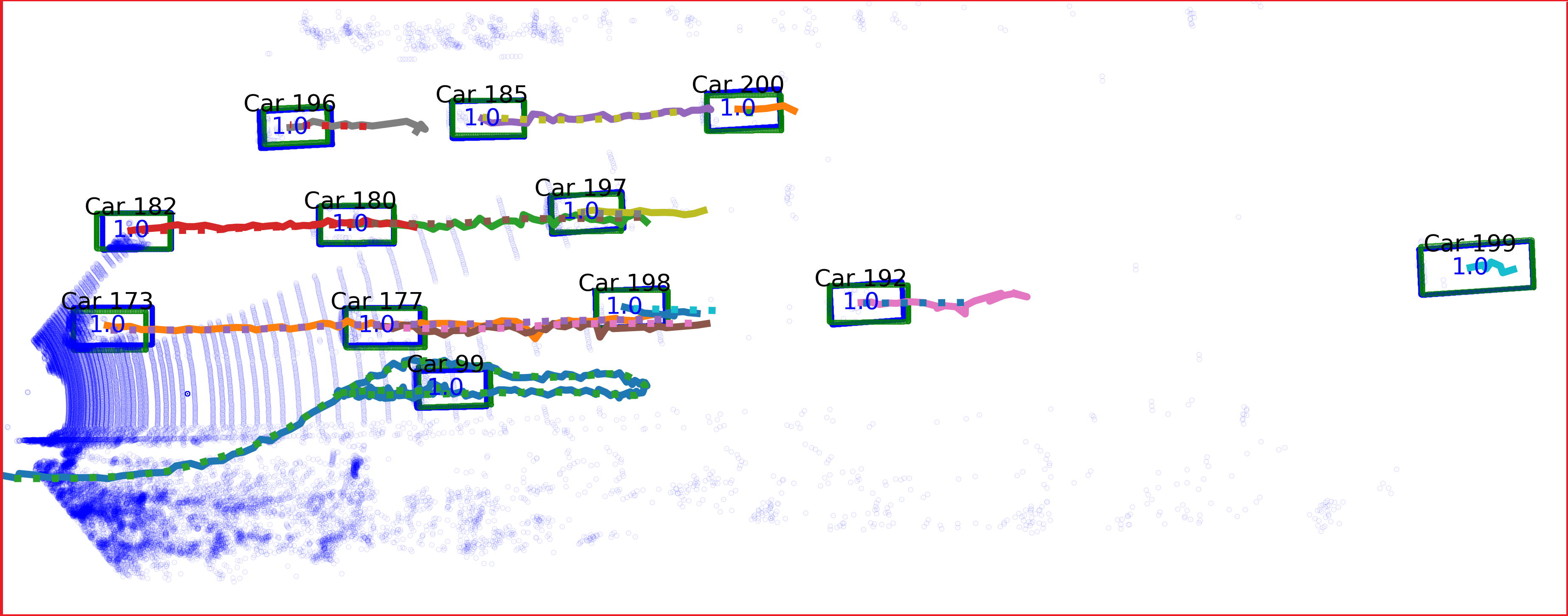}
	\caption{Birds-eye-view of our detection and tracking results. The dotted lines and green bounding boxes represent the ground-truth, and the blue boxes and solid lines represent the predicted detection and tracking results. The figure is best viewed in color.}
	\label{fig:total1}
\end{figure}

Object detection outputs bounding boxes of each object, which is the foundation of MOT. The tracking is to associate the bounding box of each object across different frames. The ID assigned to each bounding box is unique during tracking. Traditional tracking methods associate bounding boxes using motion prediction algorithms like Kalman filters, or particle filters, as well as bipartite graph matching methods like the Hungarian algorithm \cite{kuhn1955hungarian, breitenstein2009robust}. They are vulnerable to extreme motion conditions, such as sudden braking and turning.
In addition, they can be greatly degraded by unsatisfactory detection results. For example, if an object is lost in the detection, the bounding box of this object is likely to be assigned to other objects, hence leading to tracking errors.  
  
In this letter, we propose an end-to-end 3-D object detection and tracking network using only a single Lidar sensor, which could alleviate the multi-sensor issue and the limitation of the traditional methods. Sample results of our network can be found in Fig. \ref{fig:total1}. Our network takes as input the adjacent last and current point-cloud frames, and outputs the bounding boxes as well as the tracking trajectories. We perform experiments on the KITTI benchmark dataset and the results confirm the effectiveness of our design. The contributions of this work are listed as follows:

\begin{itemize}
\item We propose an end-to-end 3-D object detection and tracking network, which takes as input two adjacent raw point clouds, and outputs the predicted bounding boxes as well as the point-wise association displacements.
\item We propose a novel data association module in our network to merge the point features of the two frames and associate the corresponding features of the same object.
\item We generate the predicted bounding boxes from the point-wise data association. The predicted bounding boxes can refine the detection results.
\end{itemize}

The remainder of this letter is organized as follows. Section II reviews the related work. Section III describes our network in detail. Section IV presents the experimental results and discussions. Conclusions and future work are drawn in the last section.

\section{Related Work}

\subsection{3-D Object Detection}
The 3-D object detection is to localize and classify objects in 3-D scenes. The robust and accurate spatial information captured by a LiDAR is conducive to increasing the accuracy of detection results. In \cite{beltran2018birdnet}, the point clouds were projected  into a bird's-eye-view where 2-D CNN could be applied on the 2-D images. In \cite{li20173d},  point cloud was voxelized into grids, then high-level features could be extracted from the hand-crafted features encoded in the grids. In \cite{zhou2018voxelnet},  voxel-wise features was extracted, instead of hand-crafted pre-prepared features. In \cite{shi2019pointrcnn},  PointNet++\cite{qi2017pointnet++} was utilized as a backbone network for 3-D object detection because of its simplicity and efficiency. PointNet++ can learn features directly from raw disordered point clouds, to solve the problem of high computational cost and information loss caused by dense convolution after quantization. 


\subsection{Data Association}
Filter-based and batch-based methods are two choices for data association in multiple-object tracking.
 
Most batch-based approaches chose the 3-D objects' accurate trajectories as the output of the MOT problem\cite{frossard2018end}. Convolutional siamese networks\cite{zbontar2015computing} were broadly used for appearance feature extraction and similarity cost computation. After getting the localization results from the detection network, matching net\cite{baser2019fantrack} compared each object in adjacent frames and used learned metrics to find the corresponding pairs and produce discrete trajectories. In \cite{frossard2018end},  a scoring net and matching net were proposed, and  the tracking problem was changed to an end-to-end learning linear programming (LP) problem by designing a backpropagating subgradient descending equation through the linear program. These learning-based data-association methods were more robust and accurate than conventional methods, but the detected bounding box accuracy also limited the accuracy of the tracking or trajectory.
  
Filter-based approaches used Kalman filters \cite{wang2019pointit, bewley2016simple, wojke2017simple} and Gaussian processes \cite{hirscher2016multiple} to generate the association matrix between adjacent frames and used the Hungarian algorithm\cite{bewley2016simple} to optimize it. PointIT\cite{wang2019pointit} used  spherical images generated from 3-D point clouds to get each object's location and extended the SORT\cite{bewley2016simple} algorithm to finish the real-time tracking problem. The filter-based methods were efficient enough for real-time deployment. However, the error accumulation problem can scarcely be solved.

These existing MOT algorithms always need the  objects' precise bounding boxes in each frame of a sequence. Several optical flow and scene flow have made some achievements, not only on 2-D images\cite{dosovitskiy2015flownet} , but also on 3-D point clouds \cite{liu2019flownet3d}. If we have the correct corresponding points pairs, and the displacement in adjacent frames, the movement of each object can be estimated exactly. Our proposed  association module, which only estimates the objects and learns their point-wise displacement, can be added to any of the existing methods, and will improve the accuracy of data association.

\begin{figure*}[t]
	\centering
	\includegraphics[width=1\textwidth]{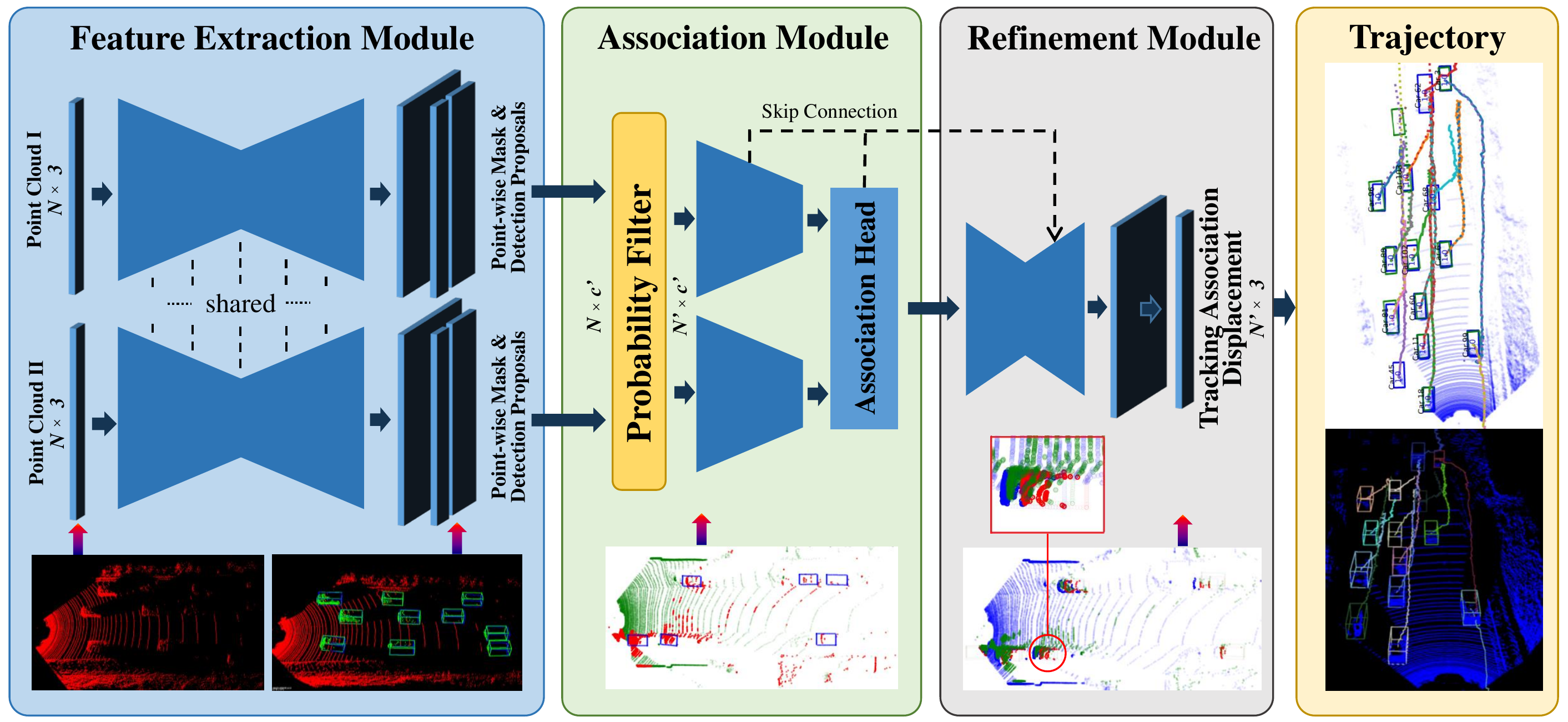}
	\caption{PointTrackNet architecture. Feature Extraction Module outputs both  point-wise mask and object bounding boxes. Association Module has a probability filter to reserve the high probability foreground points, and an association head to fuse the features of the two frames.   Refinement Module   outputs the point-wise tracking association displacements. And trajectory generator matches the same object and visualizes the bird's-eye-view and 3-D trajectories}
	\label{fig:schematic}
\end{figure*}

\section{The Proposed Network}
Our proposed network only takes as input two adjacent raw disordered point clouds, and outputs object bounding boxes and the trajectory of each object. The detection and data association are the two main components of tracking methods. We design a point-wise data association method to reduce the possible negative impacts caused by degraded object detection.


Fig. \ref{fig:schematic} shows the overview of our network architecture. The network can be split into four modules: the point-wise feature extractor (3-D object detector), the data association module, the refinement module, and the trajectory generator module. In the following, we will introduce the four components as well as the loss function.

\subsubsection{Point-wise Feature Extractor}
Given as input an $N \times 3 $ point cloud, the object detector is proposed to generate an $N \times 2$ mask and $ M $ bounding boxes, where $N$ means the number of the points and the mask is a binary 0-1 classification label to distinguish foreground and background. The point cloud feature is extracted from a backbone network. We leverage PointNet\cite{qi2017pointnet} and PointNet++\cite{qi2017pointnet++} to learn point features instead of hand-crafted features. The backbone network processes the point cloud in the intuitive Euclidean space. With the disordered point cloud as input, it learns to overcome the non-uniformity and combines multi-scale neighborhood features to produce efficient and robust features.
 
In this letter, we propose a detector with four \textit{Set Abstraction (SA) Modules} to downsample the point cloud, four \textit{Feature Propagation (FP) Modules} to upsample from fewer points to more points and two \textit{Fully Connected Layers} to finally output the mask and bounding boxes results. The SA modules downsample the lower level points $\{p^{l-1}_i\}_{i=1}^m$ to the higher level points $\{p^{l}_i\}_{i=1}^n$ where $m > n$, while a Set Upconv layer propagates them in the opposite direction. Same scale grouping (SSG) and multi-scale grouping (MSG) with farthest point sampling (FPS) \cite{eldar1997farthest} are applied in the SA module, where each layer's points are the subset of the upper layer's points. The grouping layer in the FP module constructs local region sets by finding the nearest points in the higher level points, which means increasing the number of points in the feature propagation.

Inspired by F-Pointnet \cite{qi2018frustum}, we set shape of the detector's output as $N \times (2+3+2 \times NH+4\times NS+NC)$. The first $2+3$ means the probability and center of the bounding box, $2\times NH$ means the number of classes with the residuals of the heading bins, $4\times NS$ means the number of classes with the residuals of size (bounding box length, width, and height), and $NC$ means its semantic class.

\subsubsection{Association Module}
\begin{figure}[t]
	\centering 
	\includegraphics[width=0.48\textwidth]{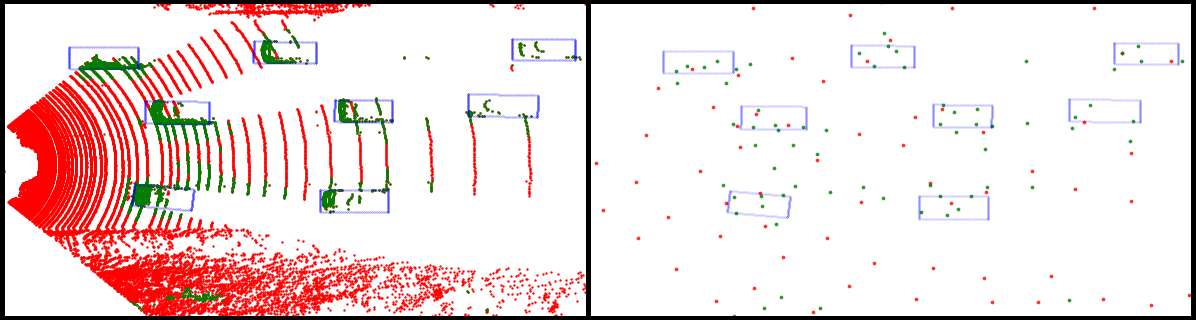}

	\caption{Comparison between original FPS sampling and filter FPS sampling.  The left is the result in TRA\_SA1 and the right is in TRA\_SA3 (detailed in Tab. \ref{tab:architecture}). The green points are the FPS sampling result after the probability filter. The red points are the original FPS sampling result. Blue boxes are the ground truth object bounding boxes. The figure is best viewed in color.}
	\label{fig:sampling}
	\vspace{-0.5cm}
\end{figure}

The association module contains a probability filter, two SA layers, and an association head. The probability filter is proposed to balance the fore-background points and decrease the computation cost. The point-wise mask in the feature extraction module can be used as the attention map to choose the salient points. We use the mask probability information to keep the top $N'$ foreground points, which provide much more useful local geometry features than most of the background points. For the reserved $N'$ points, the FPS in the sampling layers is used, the same as in the conventional methods. Fig. \ref{fig:sampling} presents the comparative results of the FPS used in the original raw point clouds with the filtered point clouds. We visualize the sampling results in TSA 1 and TSA 2 (details in Tab. \ref{tab:architecture}). We find that there are rare points belonging to the foreground if we use FPS sample after four times, especially in distant vehicles, in a wide range of environments, but our method can reserve more foreground points in the entire process. Two SA layers are connected to down-sample the filtered point cloud.

\begin{figure}[h]
	\centering 
	\includegraphics[width=0.43\textwidth]{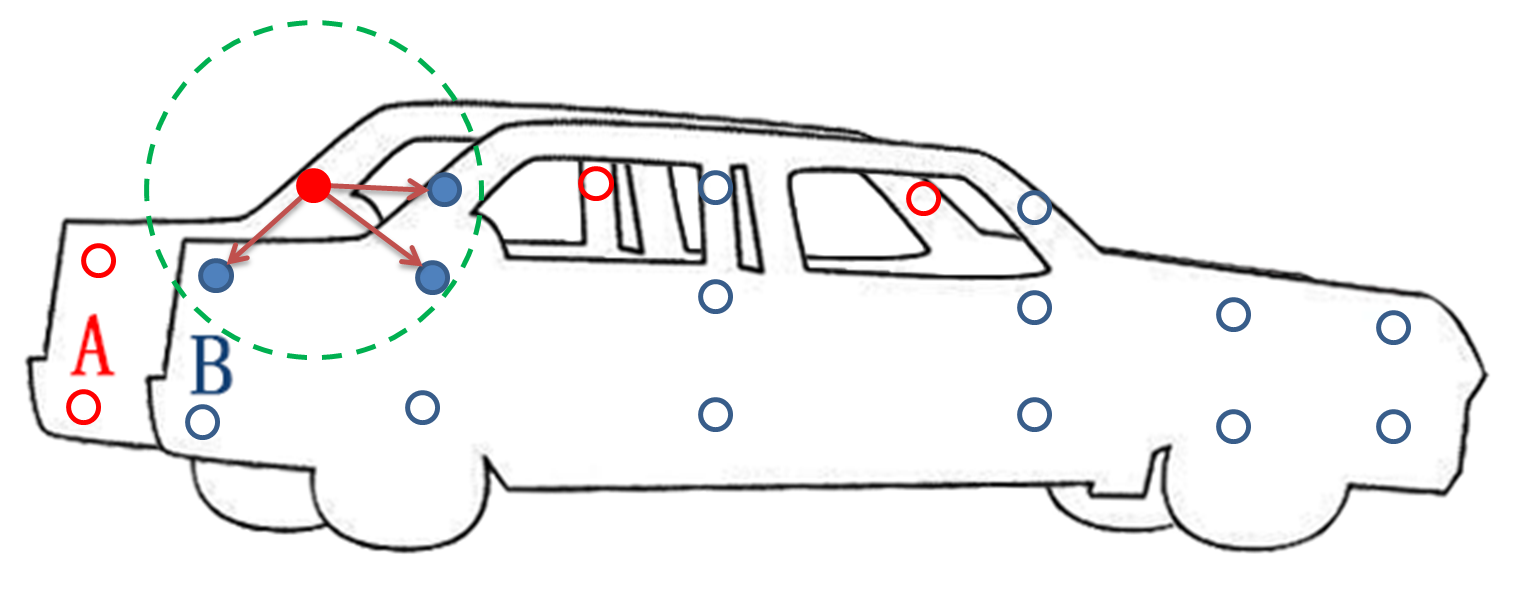} 
	\caption{Association Head visualization. Red points in frame \textbf{A} are in the past frame in time ${t-1}$, and blue points in frame \textbf{B} are in the current frame in time $t$. Solid blue points are the grouped $K$ points closest to the single chosen solid red point. We fuse the features of red solid points and blue solid points, as the embedded features. The figure is best viewed in color.}
	\label{fig:association}
	\vspace{-0.5cm}
\end{figure}

Then an association head module is proposed to merge the features of two frames. The inputs are two adjacent filtered point cloud points: $\{p_i^{t-1}=(x_i^{t-1}, y_i^{t-1}, z_i^{t-1}); p_i^t=(x_i^t, y_i^{t}, z_i^t)\}_{i=1}^{N'}$ and features$\{f_i^{t-1}; f_i^{t}\}_{i=1}^{N'},$ where $t$ ranges from 0 to the end of a sequence, $N'$ is the number of filtered points, the feature vector $f_i \in \mathbb{R}^c $.
The module learns an embedded feature $f^{'}_{i} \in \mathbb{R}^{c^{'}}$ that associates with point $ p_i^{t-1} $ in the first frame.

The visualization of the association head module is shown in  Fig. \ref{fig:association}. For every point $p^{t-1}_i$, we find the nearest $K$ points in the point set $\{p_{j}^{t}\}_{j=1}^K \subset \{p^{t}\}$, with the features $\{f_{j}^{t}\}_{j=1}^K$, and the Euclidean distance $\{d_{j}^{t}\}_{j=1}^K \subset  \mathbb{R}^3$ between $p^{t-1}_i$ with new points in the next frame. After concatenating $f^{t-1}_i$, $\{f_{j}^{t}\}_{j=1}^K$ and $\{d_{j}^{t}\}_{j=1}^K$ as embedded features, we try to use multi-layer perceptrons and element-wise max pooling to learn the point-wise association tracking displacement between the points of two adjacent frames.

\subsubsection{Refinement Module}
The refinement module is composed of one SA layer, three Set Upconv layers and two fully connected layers. For the second Set Upconv layer, we introduce a
skip connection from  TSA\_L1$_{f1}$ to concatenate with it. The detailed layer
parameters are shown in Tab. \ref{tab:architecture}.

The refinement module is proposed to refine the association features. The output of the last Set Upconv layer $\{f^{l=0}_i\}_{i=1}^{N'} \subset  \mathbb{R}^{c'}$ has the same number of points ${N'}$ as the points after the filter step.

Two FC layers are connected to calculate the final tracking results $\{t_i\}_{i=1}^{N'} \subset \mathbb{R}^3$, where $t_i$ means point-wise association displacement in two frames.
 
\subsubsection{Trajectory generator}
We use the predicted point-wise association tracking displacement to predict the movement of the bounding boxes of objects in the last frame.
The complete pseudocode in the tracking management is described in Algorithm 1.
 
\begin{algorithm}[t]
\caption{Association to Trajectory}
\LinesNumbered 
\KwIn{Detected object bounding boxes $B_{t - 1}$ and $B_t$ in time $t$; Point-wise association displacement $d_i$; LiDAR points $P_t$}

	\eIf{$t = 0$}{
		\ForEach{$B_i^{0}$}{
			initialize new trajectory and save
		}
	}{
		\ForEach{$B_i^{t-1}$}{
			find $ \forall p_m \in B_i^{t-1}$
			
			$\bar{d} = \frac{1}{N_{p_{m}}}\sum_{p_m \in B_i^{t-1}}t_m$
			
			$\widehat{Pos}_{B_i^{\tau}} = Pos_{B_i^{t}} + \bar{d}$
			
			\eIf{$max(IoU)$ for each $B_i^{t} > \tau$}
			{
				Add this pair to $Match\_Pairs$
			}{
				Add $B_i^{t-1}$ to $Unmatch\_List_{t-1}$
			}
		}
		\ForEach{$B_i^{t-1}$}{
			\If{$B_i^{t-1}$ hasn't been chosen}{
				Add $B_i^{t}$ to $Unmatch\_List_{t}$	
			}
		}	
		Use $Match\_Pairs$, $Unmatch\_List_{t-1}$ and $Unmatch\_List_{t}$ to update trajectory and save
	}

\end{algorithm}

\subsubsection{Loss Function}

The loss can be split into two parts, detection loss and  tracking loss, as (\ref{eq1}):
\begin{equation}
\label{eq1}
\mathcal{L} = \mathcal{L}_{det} + \mathcal{L}_{tracking}.
\end{equation}
Similar to \cite{qi2018frustum}, we set the detection loss as (\ref{eq2}), where $\mathcal{L}_{mask}$ means the point-wise fore-background loss, $\mathcal{L}_{box}$ means bounding box loss in (\ref{eq3}), and $\mathcal{L}_{sem\_cls}$ means the semantic class loss.
$\mathcal{L}_{mask}$  and any \textit{class}-relevant loss uses the focal loss\cite{yun2019focal} to tackle the fore-background imbalance problem, as  (\ref{eq4}):
\begin{equation}
\label{eq2}
\mathcal{L}_{det} = \mathcal{L}_{mask} + \mathcal{L}_{box} + \mathcal{L}_{sem\_cls},
\end{equation}
\begin{equation}
\label{eq3}
\mathcal{L}_{box} = \mathcal{L}_{center} + \mathcal{L}_{h\_cls} + \mathcal{L}_{h\_reg} + \mathcal{L}_{s\_cls} + \mathcal{L}_{s\_reg},
\end{equation}
\begin{equation}
\label{eq4}
\mathcal{L}_{*\_cls} = -(1-p_t)^\gamma log(p_t).
\end{equation}

We use the mask information to split the total point cloud into positive part and negative part,  where the positive means foreground and the negative means background.  Then we apply the weighted L2 Loss in the tracking-net in  (\ref{eq5}):
\begin{equation}
\label{eq5}
\mathcal{L}_{tracking} = \alpha \frac{N}{N_{pos}}L2^{pos} + \beta \frac{N}{N_{neg}}L2^{neg},
\end{equation}
where $L2^{pos}$ and $L2^{neg}$ are the L2 loss functions of positive and negative point clouds, which are distances between predicted tracking displacement with the ground truth. $N$ means the number of points in the total point cloud, $N_{pos}$ and $N_{neg}$ are number of points in each part of points.

\section{Experimental Results and Discussions}

\subsection{Dataset and Evaluation Matrices}
We evaluate our model on the 3-D object tracking benchmark dataset from KITTI\cite{geiger2012we}. 
The dataset consists of 21 training and validation sequences with publicly given ground truth labels and 29 test sequences that need to evaluate on-line. For the training and validation set, there are 7987 point-cloud frames, 636 vehicle trajectories, and 30601  individual vehicles in total.
We split the training and validation set into two parts, \textit{Seq-0000} to \textit{Seq-0015} for training and \textit{Seq-0016} to \textit{Seq-0020} for validation.

During the training process, we firstly extract the ground points, and label the points in the ground truth bounding boxes as foreground points and the others as background. Then, we compare  two objects in adjacent frames which have the same object ID, and substitute the bounding boxes' movement for  point-wise association tracking displacement.
During the testing process, we only feed two adjacent raw disordered point clouds to the network.

The CLEAR MOT metrics \cite{bernardin2006multiple} are used in our method for evaluating detection and tracking accuracy. Mostly Tracked (MT), Mostly Lost (ML), ID Switches (IDS) and FRAGmentation (FRAG) can reflect tracking orientation characteristics. A higher MT and a lower ML, IDS, and FRAG mean the tracker has improved in continuous tracking and reduced the trajectory FRAG and IDS.

\subsection{Network Details}
The detailed parameters of each layer are shown in Tab. \ref{tab:architecture}. 
Each learnable layer adopts multi-layer perceptrons with a few Linear-BatchNorm-ReLU layers parameterized by its linear layer width.
\begin{table}[]
\renewcommand\arraystretch{1.0}   
\renewcommand\tabcolsep{8pt}  
\setlength{\abovecaptionskip}{0pt} 
\setlength{\belowcaptionskip}{0pt} 
\captionsetup{justification=centering}
		\caption{\textsc{Detailed Network Configurations.}}
		\label{tab:architecture} \centering %
	\begin{tabular}{@{}ccccc@{}}
		\toprule
		& Layer & Radius & Point Num & MLP width         \\ \otoprule
		\multirow{10}{*}{DET} & SA1   & 1.0      & 2048        & {[}32,32,64{]}    \\
		& SA2   & 2.0      & 512         & {[}64,64,128{]}   \\
		& SA3   & 4.0      & 128         & {[}128,128,256{]} \\
		& SA4   & 8.0      & 32          & {[}256,256,512{]} \\
		& FP1   & 8.0      & 128         & {[}256,256{]}     \\
		& FP2   & 4.0      & 512         & {[}128,128,256{]} \\
		& FP3   & 2.0      & 2048        & {[}128,128,256{]} \\
		& FP4   & *      & N           & {[}256, 256{]}    \\
		& FC1   & *      & *           & 128               \\
		& FC2   & *      & *           & C                 \\ \otoprule
		\multirow{10}{*}{TRA}  & Pr Filter (f1, f2)   & *    & N'        & *    \\
        & SA1 (f1, f2)   & 0.5    & 2048        & {[}32,32,64{]}    \\
		& SA2 (f1, f2)   & 1.0      & 512         & {[}64,64,128{]}   \\
		& Assn. & 5.0      & 512         & {[}128,128{]}     \\
		& SA3   & 4.0      & 32          & {[}256,256{]}     \\
		& FP1   & 4.0      & 512         & {[}256,256{]}     \\
		& FP2   & 1.0      & 2048        & {[}128,256,256{]} \\
		& FP3   & *      & N'           & {[}256, 256{]}    \\
		& FC1   & *      & *           & 128               \\
		& FC2   & *      & *           & 3                \\ \bottomrule
	\end{tabular}
	\vspace{-0.2cm}
\end{table}
 
We train the detection network and tracking network separately. The detection network is trained using a batch size of 8 and the tracking  using a batch size of 6. We use an Adam\cite{kingma2014adam} optimizer for training, in which beta1=0.9,
beta2=0.999, epsilon=1e-08. The cyclical learning rates (CLR)\cite{smith2018super} is used in our architecture to lead a higher training speed. The learning rate changes linearly,  upper bound is 0.001, lower bound is 0.0001, and the cycle range is 8 epochs.

The proposed PointTrackNet is implemented on Tensorflow 1.9.0 with CUDA 9.0 and cuDNN 7.0 libraries. It is trained on an Intel 3.7GHz i7 CPU and a single GeForce GTX 1080Ti graphics card.

The number of points in point-wise point cloud network is the key factor in computation cost. We compare the computation time of two raw point cloud inputs in the different sets of input size and the probability filter size in Tab. \ref{tab:time}.
In our experiment, we set the number of input point cloud $N$ to 15000, the number of points after probability filter $N'$ to 5000, and the $K$ in Association Module to 64.

\begin{table}[t]
\renewcommand\arraystretch{1.0}   
\renewcommand\tabcolsep{14pt}  
\setlength{\abovecaptionskip}{0pt} 
\setlength{\belowcaptionskip}{0pt} 
	\captionsetup{justification=centering}
	\caption{\textsc{Analysis of Computation Time. The Unit is Second.}}
	\label{tab:time} \centering %
	\begin{center} 
		\begin{tabular}{@{}cccc@{}}
		\toprule
			K=64          & N'=5000 & N'=1000 & N'=500\\ \midrule
			Input N=15000 & 0.099    & 0.086    & 0.085   \\ 
			Input N=10000 & 0.074    & 0.060     & 0.059  \\
			Input N=5000  & 0.048    & 0.037    & 0.036  \\
		\bottomrule
		\end{tabular}
	\end{center}
	\vspace{-0.6cm}
\end{table}

\subsection{Ablation Study}
An ablation study is designed to evaluate various multi-feature fusion methods in the association head module. The association module uses the local features from two detectors to learn the weight related to the displacement of each point in the first frame. Which layer's features to choose and how to merge these features in different frames are studied in this section.

Tab. \ref{tab:ablation1} illustrates the ablation study result with various feature fusion methods.  We consider the coordinates difference between $p^{t-1}_i$ and $\{p^{t}_j\}_1^K$ as the "fundamental" association displacement. So if we use a feature-correlated fusion method, like the  cosine distance or dot product, the closer the characteristics of the two points are, the greater the value after the fusion will be. That is the reason why these two methods show advantages over the other two methods in Tab. \ref{tab:ablation1}.
\begin{table}[tp]
\renewcommand\arraystretch{1.0}   
\renewcommand\tabcolsep{9pt}  
\setlength{\abovecaptionskip}{0pt} 
\setlength{\belowcaptionskip}{0pt} 
\captionsetup{justification=centering}
\caption{\textsc{Ablation Study Results of Evaluation Matrices with Various Feature Fusion Methods. Bold Font Highlights the Best Results.}}
	\label{tab:ablation1} \centering %
\begin{tabular}{@{}ccccc@{}}
\toprule
Fusion method      & MOTA$\uparrow$&MOTP$\uparrow$&MT$\uparrow$&ML$\downarrow$    \\\midrule
elementwise poduct & 0.704          & 0.792          & 0.539         & 0.080          \\
concat             & 0.717          & 0.790          & 0.65          & 0.044 \\
cosine distance    & \textbf{0.747} & \textbf{0.811} & 0.68          & 0.047          \\
dot product        & 0.734          & 0.810          & \textbf{0.74} & \textbf{0.036} \\
\bottomrule
\end{tabular}
\end{table}

Tab. \ref{tab:ablation2} illustrates the ablation study results with different layers' features to fuse.  We choose the features from $DSA2 + DFP2$, $DSA3 + DFP3$, $filter + FC1$, $filter + FC2$ to evaluate. We can find that using the probability filter described in Sec. III.(2) is better than using the original farthest point sampling result. So using more features in FC1 is better than FC2, though it will slightly increase the computational load.

\begin{table}[tp]
\renewcommand\arraystretch{1.0}   
\renewcommand\tabcolsep{11pt}  
\setlength{\abovecaptionskip}{0pt} 
\setlength{\belowcaptionskip}{0pt} 
\captionsetup{justification=centering}
\caption{\textsc{Ablation Study Results of Evaluation Matrices with Various Layers' Features to Fuse. The Bold Font Highlights the Best Results.}}
	\label{tab:ablation2} \centering %
\begin{tabular}{@{}ccccc@{}}
\toprule
Fusion method      & MOTA$\uparrow$&MOTP$\uparrow$&MT$\uparrow$&ML$\downarrow$    \\\midrule
DSA2 + DFP2 & 0.602          & 0.800          & 0.639         & 0.062          \\
DSA3 + DFP3             & 0.634          & 0.795          & 0.612          & 0.075 \\
Filter + FC2    &0.707& \textbf{0.801} & 0.685          & 0.040          \\
Filter + FC1        & \textbf{0.739}          & 0.800          & \textbf{0.710} & \textbf{0.034} \\
\bottomrule
\end{tabular}
\vspace{-0.3cm}
\end{table}

\begin{figure}[t]
	\centering
	\includegraphics[width=0.5\textwidth]{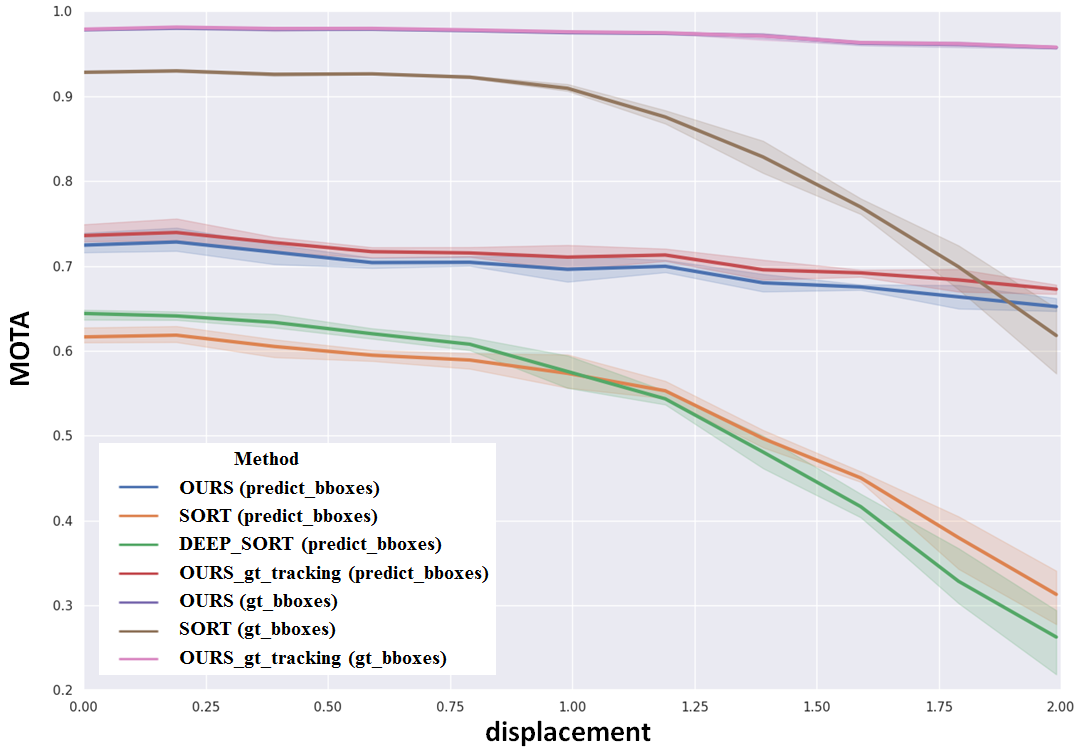}
	\caption{Comparison results with baseline trackers on KITTI validation dataset. The MOTA results of three different trackers in various manual displacement ranges from 0.0 to 2.0 (meters). '(gt\_bboxes)' means that the method uses ground truth detection result, and 'gt\_tracking' means that the method uses ground truth point-wise association tracking displacement. The figure is best viewed in color.}
	\label{fig:compare_baseline}
	\vspace{-0.3cm}
\end{figure}

\begin{figure*}[t]
	\centering
	\includegraphics[width=1\textwidth]{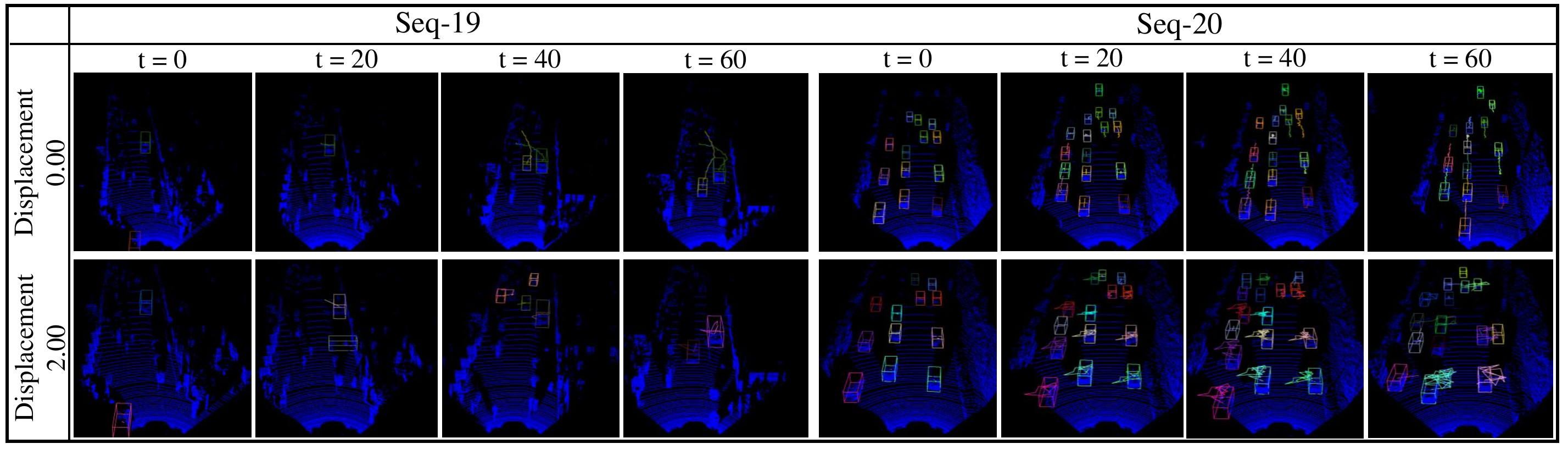}
	\caption{The qualitative results of sequences 19 and 20. The top two rows show the original
		point cloud inputs and the bottom two rows show the augmented point cloud with random displacement less than 2.0. From left to right, the same object at different time stamps is represented by the same color and the lines are trajectories. The figure is best viewed in color.}
	\label{fig:qualitative}
	\vspace{-0.5cm}
\end{figure*}

\begin{table*}[h]%
\setlength{\abovecaptionskip}{0pt} 
\setlength{\belowcaptionskip}{0pt} 
\captionsetup{justification=centering}
	\caption{\textsc{Comparative Results of Evaluation Matrices with Various Displacements for Different Trackers. The Bold Font Highlights the Best Results.}}
	\label{tab:comparison_result} \centering %
	\begin{tabular}{@{}cccccccccccccc@{}}
		\toprule
		\multicolumn{2}{c}{\multirow{2}{*}{Methods}} &  \multicolumn{6}{c}{$ Displacement=0.0 $} & \multicolumn{6}{c}{$Displacement=2.0$}\\ \cmidrule(l){3-8} \cmidrule(l){9-14}
		\multicolumn{2}{l}{}             & MOTA$\uparrow$&MOTP$\uparrow$&MT$\uparrow$&ML$\downarrow$&ID\_sw$ \downarrow$&FRAG$ \downarrow$ & MOTA$ \uparrow$&MOTP$ \uparrow$&MT$ \uparrow$&ML$ \downarrow$&ID\_sw$ \downarrow$&FRAG$ \downarrow$\\\toprule
		\multirow{3}{*}{\rotatebox{90}{\tabincell{c}{pred\\bboxes}}} & SORT &	0.611 & 0.797 & 0.728 & 0.032 & 99 & 162 & 0.357&0.807&0.032&0.184&87&205\\
		
		&DEEP\_SORT               & 0.648 & \textbf{0.805} & 0.652 & 0.054 & 87 &151 & 0.259 & 0.811 & 0.000 & 0.293 & 80  & \textbf{137}   \\
		&\textbf{PointTrackNet}            & \textbf{0.738} & 0.804 & \textbf{0.750} & \textbf{0.021} & \textbf{68} & \textbf{135} & \textbf{0.647} & \textbf{0.812} & \textbf{0.445} & \textbf{0.130} & \textbf{78}  & 144   \\\midrule
		\multirow{3}{*}{\rotatebox{90}{\tabincell{c}{gt\\bboxes}}} & SORT                     & 0.927 & 0.933 & 0.976 & 0.000 & 43 & 53  & 0.601 & 1.000 & 0.228 & 0.032 & 107 & 228  \\
		&DEEP\_SORT                 & 0.926 & 0.933 & 0.976 & 0.000 & 42 & 56  & 0.698 & 1.000 & 0.423 & 0.021 & 126 & 239  \\
		& \textbf{PointTrackNet}                     & \textbf{0.978} & \textbf{0.933} & \textbf{1.000} & \textbf{0.000} & \textbf{34} & \textbf{39}  & \textbf{0.957} & \textbf{1.000} & \textbf{0.989} & \textbf{0.000} & \textbf{29}  & \textbf{44}   \\\bottomrule
	\end{tabular}
	\vspace{-0.2cm}
\end{table*}

\begin{table*}[h]
\renewcommand\arraystretch{1.0}   
\renewcommand\tabcolsep{18pt}  
\setlength{\abovecaptionskip}{0pt} 
\setlength{\belowcaptionskip}{0pt} 
\captionsetup{justification=centering}
	\caption{\textsc{Comparative Results of Evaluation Matrices on KITTI Test Dataset. * means this method is merely LiDAR-based. The Bold Font Highlights the Best Results.}}
	\label{tab:comparison_KITTI} \centering %
	\begin{tabular}{@{}ccccccc@{}}
		\toprule
		Method               & MOTA$\uparrow$&MOTP$\uparrow$&MT$\uparrow$&ML$\downarrow$&ID\_sw$ \downarrow$&FRAG$ \downarrow$         \\\midrule
CEM\cite{milan2013continuous}       & 51.94\%  & 77.11\%  & 20.00\%  & 31.54\%  & 125 & 396  \\
ODAMOT\cite{gaidon2015online}    & 59.23\%  & 75.45\%  & 27.08\%  & 15.54\%  & 389 & 1274 \\
mbodSSP\cite{lenz2015followme}  & 72.69\%  & 78.75\%  & 48.77\%  & 8.77\%   & 114 & 858  \\
SSP\cite{lenz2015followme}      & 72.72\%  & 78.55\%  & 53.85\%  & 8.00\%   & 185 & 932  \\
RMOT\cite{yoon2015bayesian}     & 65.83\%  & 75.42\%  & 40.15\%  & 9.69\%   & 209 & 727  \\

MDP\cite{xiang2015learning}       & 76.59\%  & 82.10\%  & 52.15\%  & 13.38\%  & 130 & 387  \\
MCMOT-CPD\cite{lee2016multi} & 78.90\%  & 82.13\%  & 52.31\%  & 11.69\%  & 228 & 536 \\
		aUToTrack*\cite{burnett2019autotrack}            & \textbf{82.25\%} & 80.52\%          & \textbf{72.62\%} & \textbf{3.54\%} & 1025         & 1402         \\
		FANTrack*\cite{baser2019fantrack}             & 77.72\%          & \textbf{82.33\%} & 62.62\%          & 8.77\%          & 150          & 812          \\
		Complexer-YOLO*\cite{simon2019complexer}       & 75.70\%          & 78.46\%          & 58.00\%          & 5.08\%          & 1186         & 2092         \\
DSM*\cite{frossard2018end} &  76.15\%          & 83.42\%          & 60.00\%          & 8.31\%          & 296 & 868 \\
		PointTrackNet* (Ours) & 68.23\%          & 76.57\%          & 60.62\%          & 12.31\%         & \textbf{111} & \textbf{725} \\\bottomrule
	\end{tabular}
	\vspace{-0.3cm}
\end{table*} 

\subsection{Comparative Results}
Kalman filter is used in  SORT\cite{bewley2016simple} and DEEP SORT\cite{wojke2017simple} to update the past frame's object position. However, when the movement of the objects or the collecting vehicle is irregular, the predicted update result will lead to
a severe increase of the mismatches and switches of output trajectories.
We design an experiment in which we assign a displacement to individual object cars, which can also be seen as the movement of the vehicle that collected the dataset. We augment the dataset, that is to say, apply a point-wise displacement into the points inside the bounding boxes. 

Fig. \ref{fig:compare_baseline} shows the comparison results on the KITTI validation dataset. In the figure, we can see that with  increasing  displacement, evaluation metric MOTA of the traditional trackers SORT and DEEP SORT decreases significantly. For our method, however, using the predicted tracking to compensate for the shift distance between two frames, the evaluation results are relatively stable at a high level. The results show that our method presents great superiorities in relatively larger changing environments, as well as in the cases that the data-collecting vehicle regularly moves. The ground truth tracking assists our method to have  performance which is at its upper bound. The tiny gap between our predicted tracking with the upper bound shows that even the tracking value are not so precise, the object IoU corresponding process can provide the correct results. The detailed comparison results of all evaluation matrices with various displacements are described in Tab. \ref{tab:comparison_result}, which show overall superior performance than the baseline methods.

In Tab. \ref{tab:comparison_KITTI} we compare our results with the publicly available LiDAR-based  methods and some well-known multi-sensor tracking methods from the KITTI tracking
benchmark. Our approach shows the competitive results with the state-of-the-art. We can see that in terms of some of the metrics, like IDS and FRAG, ours outperforms all the other methods. These two metrics are more related to the tracking performance, which means that despite low accuracy in the object detection, our network can still obtain   better results on the tracking. Furthermore, the detection in our end-to-end framework can be replaced by other point-wise feature extractors, to improve the tracking accuracy.

\subsection{Qualitative Evaluation}

Fig. \ref{fig:qualitative} displays sample qualitative results by running our tracker on the KITTI tracking validation sequences. We compare the tracking results on a  multi-car scenario (sequence
20) with a multi-obstacle scenario (sequence 19), and different
manually added displacements of each object to imitate the
movement of the LiDAR. We visualize the trajectories with
 predicted bounding boxes from 3-D
perspective. The bounding box and trajectory of each object share the same color. When displacement
is a random value less than 2.0, the first car, in the navy color,
is lost after 10 frames. Because our detector lost the car in
frames 7 to frame 9, the huge displacement of that car confuses the
tracker. However, the result using the original point cloud
as input shows that our tracker performs fairly well, even
when the detector lost the target car. The tracking
association information helps the tracker to reduce the impact
of  the short term disappearance of  objects. Sequence 19 with
displacement equal to 2.0 can be seen as a failure sample. The
reason could be that the object car is far away from the LiDAR so that
the feature is not integrated enough. In addition, the large number of  obstacles in front of the car could
result in serious occlusions. Increasing the training sample and
improving the detector performance could be conducive to improving results.

\section{Conclusions}
 We proposed here a novel end-to-end LiDAR-based network for 3-D object detection and tracking. The experimental results demonstrate that our network shows competitive results over the state-of-the-arts. Due to the diversity of detection approaches, we believe that the point-wise feature extractor can be replaced with more powerful one in the future to improve the overall tracking performance. We will also verify the tracking network's feasibility based on the voxel-based detector or 3-D to 2-D detector.







\bibliographystyle{IEEEtran}

\bibliography{bibtex}

\end{document}